\documentclass[conference]{IEEEtran}
\IEEEoverridecommandlockouts
\usepackage{cite}
\usepackage{amsmath,amssymb,amsfonts}
\usepackage{algorithmic}
\usepackage{graphicx}
\usepackage{textcomp}
\usepackage{xcolor}
\def\BibTeX{{\rm B\kern-.05em{\sc i\kern-.025em b}\kern-.08em
    T\kern-.1667em\lower.7ex\hbox{E}\kern-.125emX}}

\usepackage{microtype}
\usepackage{subcaption}
\usepackage{acronym}
\usepackage{hyperref}
\usepackage{graphicx}
\usepackage{caption}
\usepackage{enumitem}
\usepackage{amsmath}
\usepackage{mathtools, cuted}
\usepackage{lipsum, color}
\usepackage[ruled,vlined]{algorithm2e}
\usepackage{url}
\usepackage[]{siunitx}
\usepackage{epigraph}
\usepackage{xcolor}
\usepackage{mdframed}
\usepackage{colortbl}
\usepackage{verbatim}
\usepackage{booktabs}
\usepackage{tabularx}
	\newcolumntype{L}[1]{>{\raggedright\arraybackslash}p{#1}} 
	\newcolumntype{C}[1]{>{\centering\arraybackslash}p{#1}} 
	\newcolumntype{R}[1]{>{\raggedleft\arraybackslash}p{#1}} 
\usepackage{multirow}
\usepackage{pifont}
\usepackage{hhline}
\usepackage{listings}
\usepackage[autolanguage]{numprint}

\DeclareMathOperator*{\argmin}{arg\,min}
\newcommand{\abs}[1]{\lvert#1\rvert}

{}
{}
{}

\newacro{it}[IT]{Information Technology}
\newacro{iot}[IoT]{Internet of Things}
\newacro{iiot}[IIoT]{Industrial Internet of Things}
\newacro{manet}[MANET]{Mobile Ad Hoc Network}
\newacro{pin}[PIN]{Personal Identification Number}
\newacro{otp}[OTP]{One Time Password}
\newacro{ip}[IP]{Internet Protocol}
\newacro{mac}[MAC]{Mandatory Access Control}
\newacro{dac}[DAC]{Discretionary Access Control}
\newacro{rbac}[RBAC]{Role-Based Access Control}	
\newacro{radius}[RADIUS]{Remote Authentication Dial-In User Service}
\newacro{rfid}[RFID]{Radio Frequency IDentification}
\newacro{scada}[SCADA]{Supervisory Control And Data Acquisition}
\newacro{cots}[COTS]{Commercial Off The Shelve}
\newacro{osi}[OSI]{Open System Interconnection}
\newacro{tcp}[TCP]{Transmission Control Protocol}
\newacro{udp}[UDP]{User Datagram Protocol}
\newacro{ip}[IP]{Internet Protocol}
\newacro{plc}[PLC]{Programmable Logic Controller}
\newacro{ai}[AI]{Artificial Intelligence}
\newacro{ids}[IDS]{Intrusion Detection System}
\newacro{it}[IT]{Information Technology}
\newacro{ua}[UA]{Unified Architecture}
\newacro{cps}[CPS]{Cyber Physical System}
\newacro{rtu}[RTU]{Remote Terminal Unit}
\newacro{mtu}[MTU]{Master Terminal Unit}
\newacro{mse}[MSE]{Mean Squared Error}
\newacro{siem}[SIEM]{Security Event and Information Management}
\newacro{lstm}[\textit{LSTM}]{\textit{Long Short Term Memory}}
\newacro{svm}[\textit{SVM}]{\textit{Support Vector Machines}}
\newacro{roc}[\textit{ROC}]{\textit{Receiver Operating Characteristic}}
\newacro{tpr}[\textit{TPR}]{\textit{True Positive Rate}}
\newacro{fpr}[\textit{FPR}]{\textit{False Positive Rate}}
\newacro{arima}[\textit{ARIMA}]{\textit{Autoregressive Integrated Moving Average}}
\newacro{arma}[\textit{ARMA}]{\textit{Autoregressive Moving Average}}
\newacro{sarima}[\textit{SARIMA}]{\textit{Seasonal Autoregressive Integrated Moving Average}}
\newacro{mass}[\textit{MASS}]{\textit{Mueen's Algorithm for Similarity Search}}

\begin{document}

\title{Time is of the Essence:\\Machine Learning-based Intrusion Detection in Industrial Time Series Data
\thanks{Please cite as: S. Duque Anton, L. Ahrens, D. Fraunholz, and H. D. Schotten, "Time is of the Essence: Machine Learning-based Intrusion Detection in Industrial Time Series Data," in \textit{2018 IEEE International Conference on Data Mining Workshops (ICDMW)}, November 2018}
}

\author{\IEEEauthorblockN{Simon Duque Anton, Lia Ahrens, Daniel Fraunholz, and Hans D. Schotten }
\IEEEauthorblockA{\textit{Intelligent Networks Research Group} \\
\textit{German Research Center for Artificial Intelligence} \\
Kaiserslautern, Germany\\
\{simon, lia, daniel, hans\_dieter\}.\{duque\_anton, ahrens, fraunholz, schotten\}@dfki.de}
}

\maketitle

\begin{abstract}
The Industrial Internet of Things drastically increases connectivity of devices in industrial applications.
In addition to the benefits in efficiency,
scalability and ease of use,
this creates novel attack surfaces.
Historically,
industrial networks and protocols do not contain means of security,
such as authentication and encryption,
that are made necessary by this development.
Thus,
industrial IT-security is needed.
In this work,
emulated industrial network data is transformed into a time series  and analysed with three different algorithms.
The data contains labeled attacks,
so the performance can be evaluated.
\textit{Matrix Profiles} perform well with almost no parameterisation needed.
\textit{Seasonal Autoregressive Integrated Moving Average} performs well in the presence of noise,
requiring parameterisation effort.
\textit{Long Short Term Memory}-based neural networks perform mediocre while requiring a high training- and parameterisation effort.
\end{abstract}


\section{Introduction}
\label{sec:intro}
Over the last four and a half decades,
automation and industrial control have been ever changing.
At the beginning of industrial automation,
hardwired connections between control units and sensors or actuators respectively lacked flexibility and adaptability.
This led to the rise of \acp{plc},
allowing for a more versatile interconnection of production units.
Today,
the introduction of the \ac{iot} into production,
the so-called \ac{iiot},
is emerging.
The goal is to increase flexibility and reduce set up and configuration times,
as well as to decrease cost and effort needed.
Interconnectivity of devices among different network areas is part of this change of paradigm,
as well as connectivity through trust boundaries and over public internet.
With all the benefits arising from the \ac{iiot},
however,
risks are evolving as well.
Over the last two decades,
an increase in attacks on energy and automation systems has been noted~\cite{Duque_Anton.2017a}.
Many infamous examples have been discovered,
such as \textit{StuxNet},
\textit{Industroyer} and \textit{Black Energy}.
Unfortunately,
industrial \ac{it} security has not evolved as fast as the \ac{iiot}.
While home and office appliances always had to combat attacks,
industrial applications have been deemed secure due to two properties~\cite{Igure.2006}:
First,
the industrial control,
or \ac{scada},
system are physically separated from public networks.
And second,
the network properties are too unique for an attacker to effectively exploit them in case the perimeter was broken.
The first property obviously falls to the paradigm of interconnectivity.
As for the second one,
\ac{cots} products and standardised hard- and software have been introduced to industry as they have been into the consumer \ac{iot} market,
decreasing customisation and configuration effort and increasing efficiency.
This,
however,
enlarges attack surfaces and makes the implementation of reusable exploits possible.
Thus,
industrial \ac{it} security has become a major issue over the past years.
Unlike home and office intrusion detection,
however,
security solutions are not as mature.
Partly,
this is due to the scarcity of data to test industrial \ac{ids} applications on.
As the communication patterns of industrial systems differ from home and office-based network traffic,
the same \ac{ids} cannot be adapted for industrial application.
Due to the characteristic nature of industrial network communication,
time series analysis lends itself readily to the application.
Time series analysis has been employed in many domains so far,
e.g. in classic intrusion detection. 
In this work,
three time series anomaly detection algorithms,
\textit{Matrix Profiles},
\ac{sarima}- and \ac{lstm}-approach are evaluated on an industrial data set based on the \textit{Modbus/TCP} protocol,
a common and open source communication protocol for industrial applications.  \\ \par
The remainder of this work is structured as follows:
Section~\ref{sec:ad_for_ts} gives an an overview of time series-based intrusion detection,
as well as industrial intrusion detection.
After that,
the data set is introduced in Section~\ref{sec:ds_for_ad}.
In Sections~\ref{sec:mp},~\ref{sec:sarima} and~\ref{sec:lstm},
three time series-based algorithms for anomaly detection are applied to the data set and the results are discussed respectively.
This work is concluded in Section~\ref{sec:conclusion}.

\section{Related Work}
\label{sec:ad_for_ts}
\ac{scada} has been identified as a likely and promising target for cyber attackers~\cite{Zhu.2011, Mitchell.2014}.
Temporal properties as a feature for intrusion detection have been well-researched~\cite{Gupta.2014, Sperotto.2010}.
Different works discuss wavelet analysis of network data represented as a time series~\cite{Lu.2009, Barford.2002}.
Furthermore,
Wiener Filtering with \ac{arma} modeling is evaluated by \textit{Celenk et al.}~\cite{Celenk.2008}.
\textit{K-means clustering} of time series data is performed~\cite{Muenz.2007},
as well as statistical analysis of temporal distribution~\cite{Kim.2004, Sperotto.2008}.
Furthermore,
neural networks have been applied to industrial network intrusion detection.
\textit{Filonov et al.} propose \ac{lstm}-based intrusion detection on a synthetic data set that has been generated in their work~\cite{Filonov.2016}.
\textit{Lin et al.} analyse the time distribution of synthetic and real \ac{scada} network data and identify deviations~\cite{Lin.2017}. 
\textit{Linda et al.} create an intrusion detection system based on neural networks~\cite{Linda.2009}.
In addition to neural networks,
other techniques have been employed to detect attacks in industrial networks.
State-based intrusion detection was evaluated by \textit{Goldenberg and Wool}~\cite{Goldenberg.2013},
\textit{Fovino et al.}~\cite{Fovino.2010} and \textit{Carcano et al}~\cite{Carcano.2009}.
Additionally, 
rule-, signature- and multiattribute-based intrusion detection systems have been evaluated~\cite{Yang.2014, Lin.2013, Verba.2008}.
Since \ac{siem}-systems are increasingly relevant for industrial intrusion detection,
\textit{Oman et al.} proposed a method to integrate intrusion detection into \ac{siem}-systems for \ac{scada} networks~\cite{Oman.2008}.
There are many works addressing the application of neural networks to intrusion detection for home and office networks~\cite{Staudemeyer.2015, Bontemps.2016}.
Unfortunately,
these works lack standardised data sets for evaluation.
Often, 
the \textit{KDD Cup'99} data set~\cite{KDD.1999} that has been proven to contain artifacts that lead to overfitting~\cite{Tavallaee.2009}  is used.
The data set we are analysing in this work~\cite{Lemay.2016} has been analysed on packet level by \textit{Duque Anton et al.}~\cite{Duque_Anton.2018}.

\section{Time Series Data Set for Intrusion Detection}
\label{sec:ds_for_ad}
\textit{Modbus} is  a communication protocol for industrial applications.
It was developed by \textit{Gould Electronics Inc.} that is now owned by \textit{Schneider Electric}~\cite{Schneider-Electric.2017}.
\textit{Modbus} follows a Master/Slave concept.
It has become a de-facto standard in industrial communication~\cite{Drury.2009}.
There are different flavours of the \textit{Modbus}-protocol that are listed in Table~\ref{tab:modbus_flavours}.
\begin{table}[h!]
\renewcommand{\arraystretch}{1.3}
\caption{Flavours of \textit{Modbus}}
\label{tab:modbus_flavours}
\centering
\begin{tabular}{l l}
\toprule
\textbf{Version} & \textbf{Version} \\
\textit{Modbus RTU} & Serial connection\\
\textit{Modbus ASCII} & ASCII-encoded serial connection\\
\textit{Modbus/TCP} & TCP/IP-based communication\\
\textit{Modbus over TCP/IP} & TCP/IP-based communication with checksum\\
\bottomrule
\end{tabular}
\end{table}
Especially the solutions based on the TCP/IP stack are widely used,
as the corresponding hardware is easily available.
This helps to reduce cost and implementation effort. \\ \par
\textit{Lemay and Fernandez} created a batch of \textit{Modbus/TCP} data sets of an emulated industrial application~\cite{Lemay.2016}.
They implemented a physical simulation model of electronic circuit breakers.
This physical model was connected via a \textit{Modbus/TCP} connection.
It consisted of three to twelve software \acp{plc} that were queried periodically by one or two \acp{mtu}.
Furthermore,
aperiodic user behaviour was introduced as queries.
After the traffic was recorded,
exploits were introduced.
These exploits were generated with the penetration testing tool \textit{metasploit}~\cite{metasploit.} and are based on the TCP/IP layer.
Unfortunately,
no \textit{Modbus} protocol-specific exploits have been performed.
However,
\textit{Lemay and Fernandez} proposed a different batch of data where they used the lowest bits in the \textit{Modbus} payload as a covert channel~\cite{Lemay.2016}. \\ \par
In this work,
three different data sets have been evaluated:
\textit{ds1} is called ``Moving\_two\_files\_Modbus\_6RTU",
\textit{ds2} is called ``Send\_a\_fake\_command\_Modbus\_6RTU\_with\_operate" and \textit{ds3} is called ``CnC\_uploading\_exe\_modbus\_6RTU\_with\_operate" in~\cite{Lemay.2016}.
These data sets and their characteristics are listed in Table~\ref{tab:data_sets}.
\begin{table}[h!]
\renewcommand{\arraystretch}{1.3}
\caption{Characteristics of Analysed Data Sets}
\label{tab:data_sets}
\centering
\begin{tabular}{l  r r r r}
\toprule
\textbf{Name} & \textbf{\#  of packets} & \textbf{Length (s)} & \textbf{\# of mal. packets} & \textbf{\# of attacks} \\
\textit{ds1} & \numprint{3319} &  \numprint{190} & 75 & 4\\
\textit{ds2} & \numprint{11166} & \numprint{670} & 10 & 1\\
\textit{ds3} & \numprint{1426} & \numprint{70} & 121 & 2\\
\bottomrule
\end{tabular}
\end{table}
In this table,
the number of packets in total is listed,
the length of communication,
as well as the number of malicious packets,
the number of attacks defined as sequences of malicious packets.
All data sets have polling intervals of 10 seconds.
In contrast to \textit{ds2} and \textit{ds3},
\textit{ds1} does not contain human interaction. \\ \par
In this work,
each second of network traffic has been aggregated as one data point,
also called event.
Features that are collected and evaluated are,
among others,
the number of protocols detected,
the number of packets and bytes,
the flags and function codes in a one-hot encoding.
Due to preliminary analysis,
three features are employed in the anomaly detection:
the number of packets per second,
the number of port pairs per second and the number of IP pairs per second.

\section{Matrix Profiles}
\label{sec:mp}
\textit{Matrix Profiles} are employed in order to detect time series-based anomalies in the data sets in this section.
At first,
the algorithm is introduced in Subsection~\ref{ssec:intro_mp}.
After that,
the three data sets introduced above are evaluated in the following subsections.
Finally,
the performance of \textit{Matrix Profiles} is discussed in Subsection~\ref{ssec:discussion_mp}.

\subsection{Introduction to Matrix Profiles}
\label{ssec:intro_mp}
The \textit{Matrix Profile} algorithm is a method to calculate similarities in time series.
It has been introduced by \textit{Yeh et al.} in 2016~\cite{Yeh.2016a}.
A sequence from a time series is compared to every other sequence of the same length within the time series.
The distances are calculated and stored.
This distance is a metric for similarity.
If a sequence has a low minimal distance,
a sequence with a related characteristic is present in the time series.
If the minimal distance of a sequence is relatively high,
it is unique in the time series.
This property is suited to find outliers that can indicate attacks.
The calculation of the z-normalised distance is described in (\ref{eq:z_norm_dist}).

\begin{equation}\label{eq:z_norm_dist}
\begin{split}
d(x,y) = \sqrt{\sum_{i=1}^{m}{(\hat{x}_{i} - \hat{y}_{i})}^2} \\
\hat{x}_{i} = \frac{x_{i} - \mu_{x}}{\sigma_{x}},\quad \hat{y}_{i} = \frac{y_{i} - \mu_{y}}{\sigma_{y}}
\end{split}
\end{equation}
By employing \textit{Pearson's Correlation Coefficient}~\cite{Benesty.2009} as shown in (\ref{eq:pearson})
\begin{equation}\label{eq:pearson}
\begin{split}
corr(x,y) &= \frac{E((x - \mu_x)(y-\mu_y))}{\sigma_x \sigma_y} \\
& = \frac{\sum^{m}_{i=1}x_i y_i - m \mu_x \mu_y}{m \sigma_x \sigma_y},
\end{split}
\end{equation}
where
\begin{equation}\label{eq:mu}
\begin{split}
\mu_x = \frac{\sum_{i=1}^{m} x_i}{m}, \quad \mu_y = \frac{\sum_{i=1}^{m} y_i}{m}
\end{split}
\end{equation}
and
\begin{equation}\label{eq:sigma}
\begin{split}
\sigma_{x}^{2} = \frac{\sum_{i=1}^{m} x_{i}^{2}}{m} - \mu_{x}^{2}, \quad \sigma_{y}^{2} = \frac{\sum_{i=1}^{m} y_{i}^{2}}{m} - \mu_{y}^{2}.
\end{split}
\end{equation}
By relating this with the Euclidean distance as shown in (\ref{eq:relation})~\cite{Mueen.2010},
\begin{equation}\label{eq:relation}
\begin{split}
d(x,y) = \sqrt{2m(1-corr(x,y))}
\end{split}
\end{equation}
the working formula for distance calculation is performed as described in (\ref{eq:working_formular_dist}).
\begin{equation}\label{eq:working_formular_dist}
\begin{split}
d(x, y) = \sqrt{2m\bigg(1-\frac{\sum_{i=1}^{m} x_{i} y_{i} - m \mu_{x} \mu_{y}}{m \sigma_{x} \sigma_{y}}\bigg)}
\end{split}
\end{equation}
$x$ and $y$  are time series,
$\mu$ is the respective mean and $\sigma$ the respective standard deviation.
$m$ is the length of a sequence.
In this work,
$m$ has been set to $10$.
Each data point represents the aggregated information of one second,
the polling interval is $10$ seconds.
\textit{Matrix Profiles},
however,
are robust to changes in $m$,
adaption of this parameter led to similar results as the ones discussed in the following subsections.
There are several more efficient implementations of the distance calculation available such as \ac{mass}~\cite{Mueen.2017}.\\ \par
To obtain the \textit{Matrix Profile},
each windowed sequence of length $m$ is compared to each other sequence of length $m$ in the time series.
An interval of $\frac{m}{2}$ before and after the start of the sequence under observation is excluded,
as this would result in a trivial match.
A sequence $x_{i}$ has a distance of $0$ from itself.
After the distances are calculated,
they are stored in a matrix.
The minimum of of each column is stored, 
indicating the minimal distance of the given sequence from any other sequence.
A slight change has been made to the algorithm as proposed in~\cite{Yeh.2016a}:
A windowed sequence is only compared to sequences that have already occurred,
meaning only the distances of $x_{i}$ from all other $x_{j}$ with $j \in \{0,...,(i-\frac{m}{2})\}$ are calculated.
In order to calculate the distances for early time points as well,
two periods of 20 seconds' duration extracted from the end of \textit{ds1} were inserted into the beginning of all data sets.
These two periods do not contain malicious or manual activity and can be considered as a training data set,
where reference patterns are stored so as to prevent the first events of the investigated data from being penalised due to their early occurrence.
A possible extension for \textit{Matrix Profiles} could be the counting of similar instances in a data set.
With the algorithm employed,
each data set is checked with respect to having occurred before.
Checking additionally how often it has occurred so far could help distinguish outliers that are seldom in comparison to regular events,
e.g. in large traffic collections.\\ \par
As the concept of distance is not easily mappable to formal metrics for classifier quality,
no classic metric for classifier quality is employed in evaluating \textit{Matrix Profiles}.
Due to the length of the sliding window, 
the raise in distance is longer than the attack itself.
This would falsify a metric in creating a large amount of false positives.
Instead,
a perfect threshold is calculated in a fashion that it is minimal,
while still able to identify every attack.

\subsection{Evaluation of \textit{ds1}}
\begin{figure}[!ht]
\centering
\includegraphics[width=0.5\textwidth]{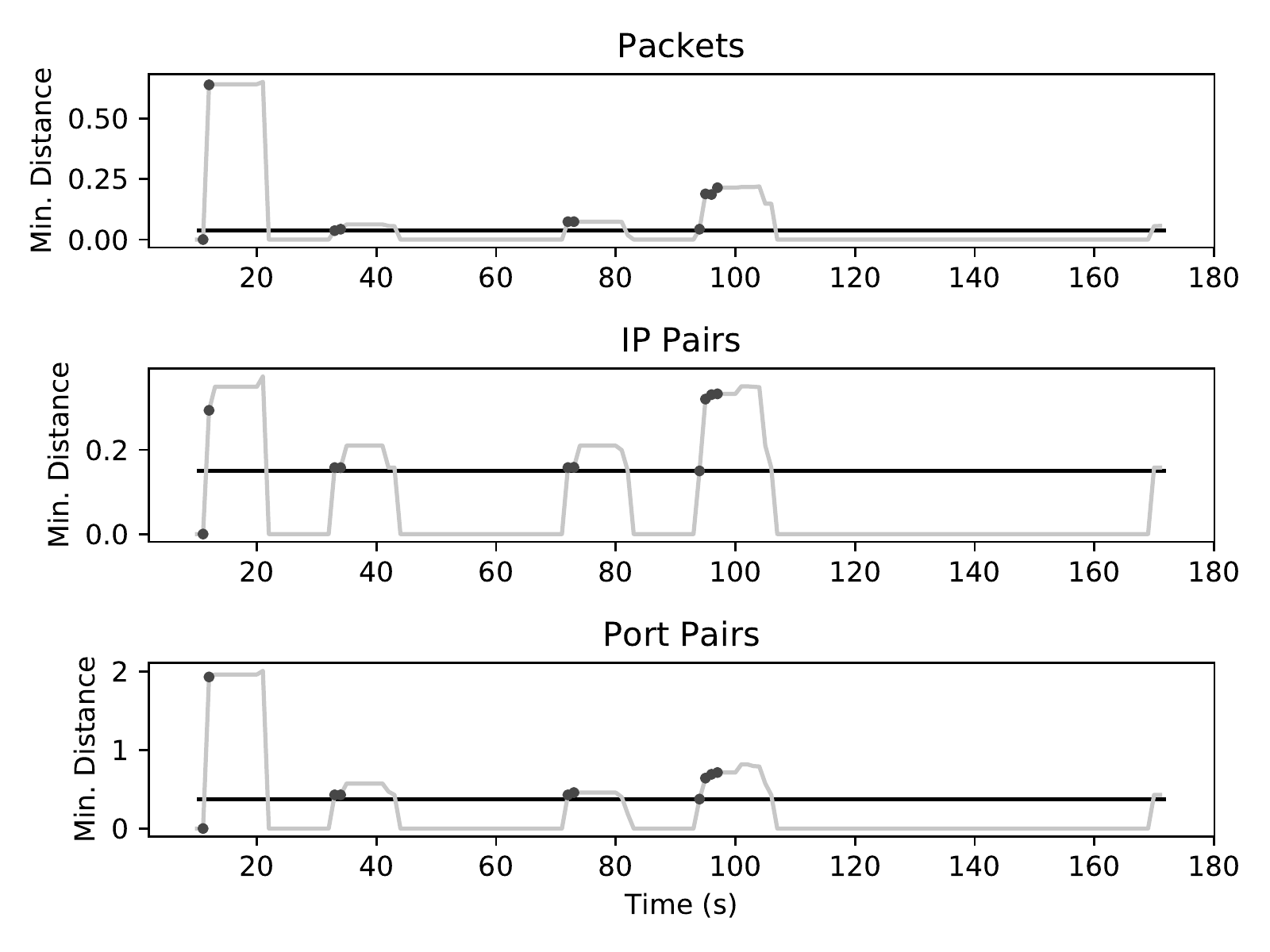}
\caption{\textit{Matrix Profile} for \textit{ds1}}
\label{fig:ds1_mp}
\end{figure}

The \textit{Matrix Profile} of \textit{ds1} is depicted in Figure~\ref{fig:ds1_mp}.
The curve describes the minimal distance of a sequence to any other,
previously occuring,
sequence.
As shown in Table~\ref{tab:data_sets},
\textit{ds1} contains four attacks in total,
depicted by the series of dark gray dots.
The attacks can be clearly distinguished by comparing the distance values to an ideal threshold,
which is displayed by a solid line.
Any feature is capable of indicating the attacks.
The four increases in distance map to the beginning of the attacks.
Only the first attack is detected with one second delay.
This is due to the fact that it falls into a polling request perfectly,
hardly altering the expected behaviour.
Its second event,
however,
clearly indicates an attack.
As \textit{Matrix Profiles} employ a sequence length of $m$,
10 in this case,
the distance value is raised for longer than the attack duration.
This could be avoided by omiting anomalous values in the calculation of distance.
Furthermore,
it is shown that \textit{Matrix Profiles} are capable of detecting attacks during the first second in which they occur,
which is an important property for in-time detection.
The increase of distance on the very end of the curves in Figure~\ref{fig:ds1_mp} is an artifact due to the data formatting.

\subsection{Evaluation of \textit{ds2}}
\begin{figure}[!ht]
\centering
\includegraphics[width=0.5\textwidth]{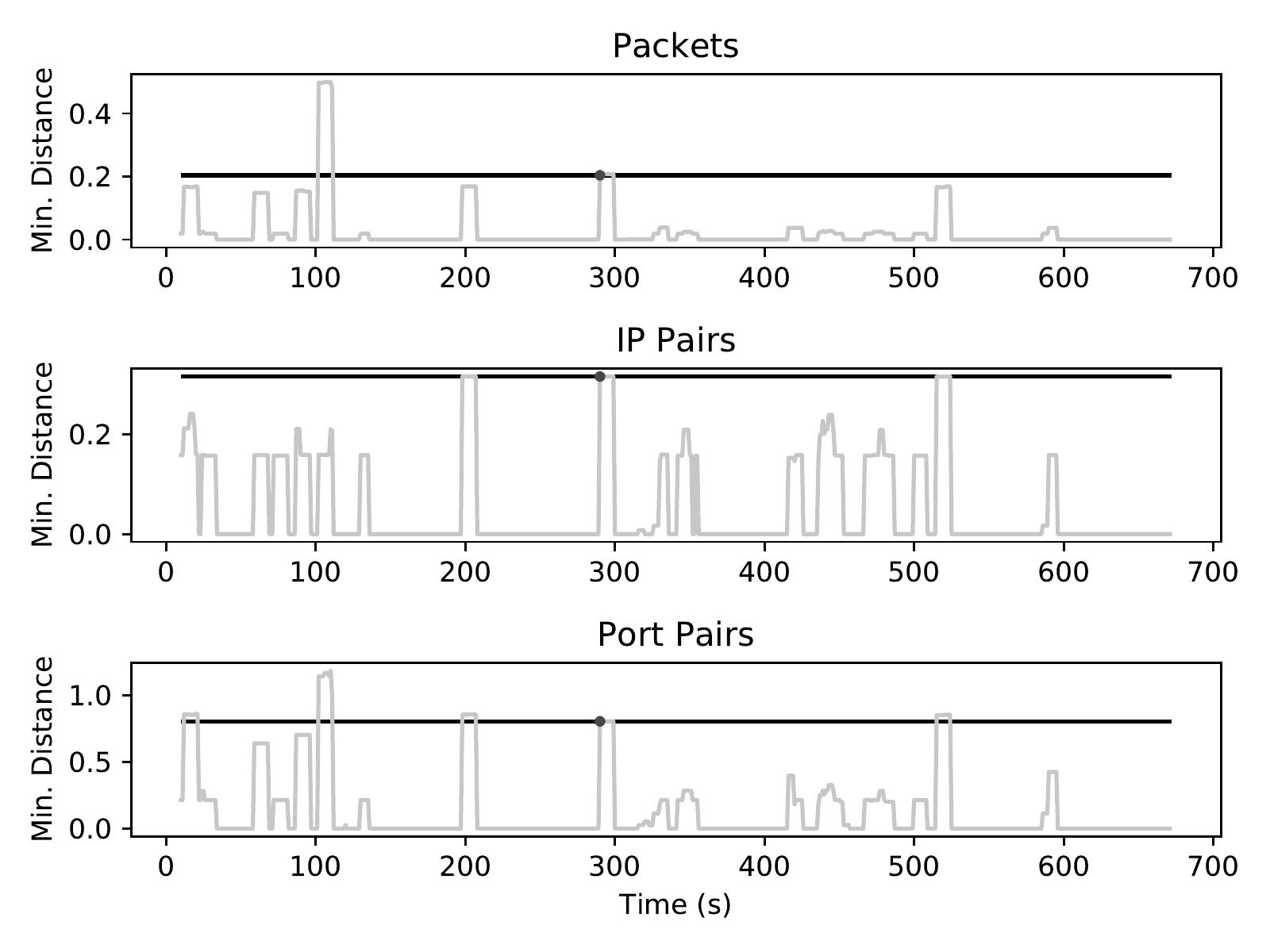}
\caption{\textit{Matrix Profile} for \textit{ds2}}
\label{fig:ds2_mp}
\end{figure}

As shown in Figure~\ref{fig:ds2_mp},
\textit{ds2} contains a lot of aperiodic traffic indicating anomalous events.
Only one of them is malicious,
marked by the dark gray dot.
In this case,
the number of packets per second is the most reliable indicator for attacks.
It would generate only one false positive if an appropriate threshold was used,
as indicated by the solid line.
The other two features create more false positives.
Despite the noise, 
a relatively good detection of the attack is possible.
Furthermore,
detecting anomalies that are not attacks is always an issue in intrusion detection.
For example,
introducing context could help in categorising singular events as anomalous but non-malicious~\cite{Duque_Anton.2017c}.

\subsection{Evaluation of \textit{ds3}}
\begin{figure}[!ht]
\centering
\includegraphics[width=0.5\textwidth]{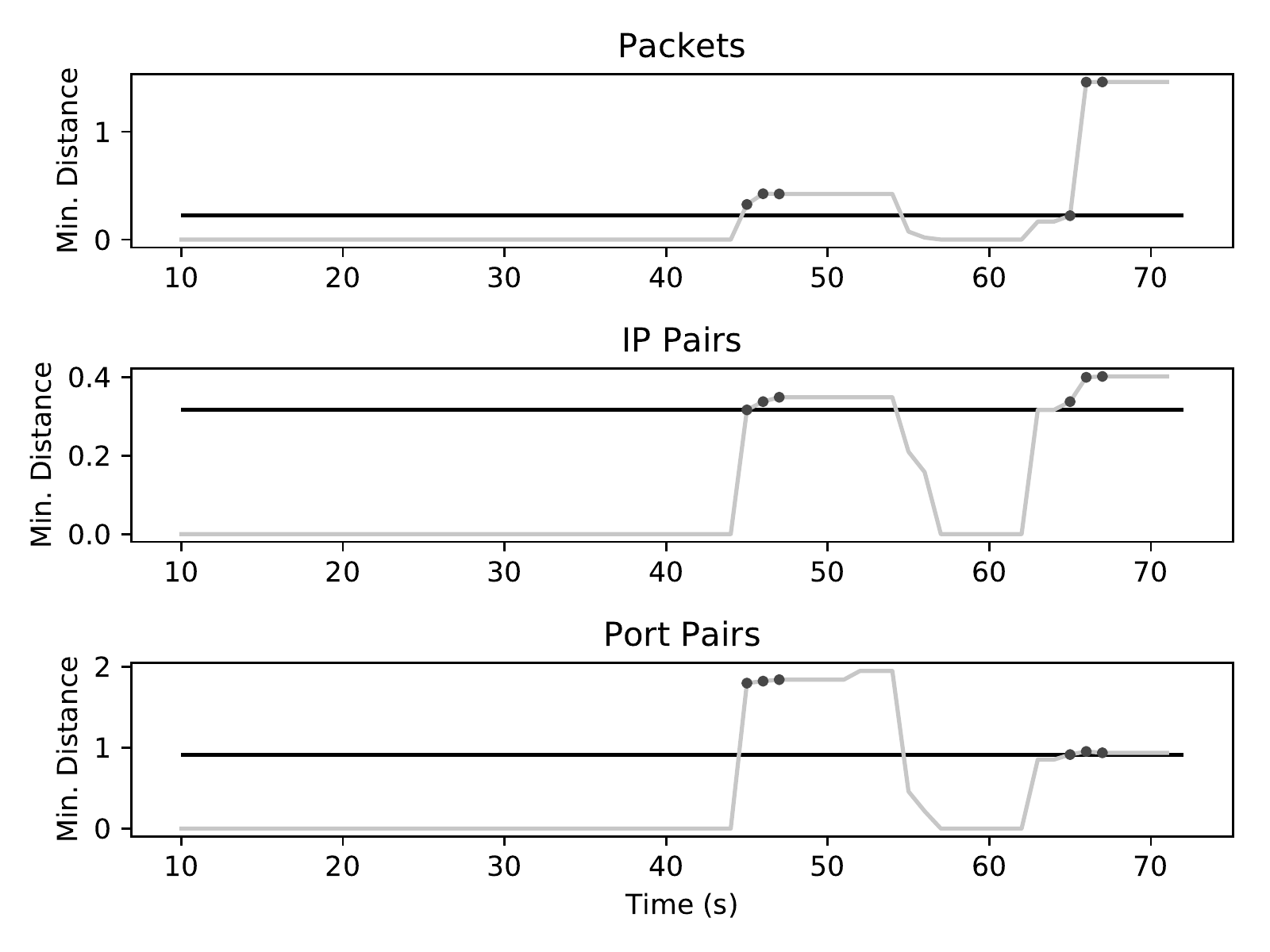}
\caption{\textit{Matrix Profile} for \textit{ds3}}
\label{fig:ds3_mp}
\end{figure}

Considering the fact that \textit{ds3} contains manual,
aperiodic operations,
the \textit{Matrix Profile} approach works exceedingly well.
The distances are shown in Figure~\ref{fig:ds3_mp}.
Especially the numbers of packets and port pairs are capable of identifying attacks that are indicated by dark gray dots,
if the perfect threshold,
illustrated by the solid line,
is employed.
As discussed before, 
the rise in distance has a longer duration due to the length of the sliding window and the influence of anomalous events on the window.
Still, 
\textit{Matrix Profiles} are relatively robust to this,
and the location of an attack can be detected easily as it is the first instance of a raised distance.

\subsection{Discussion}
\label{ssec:discussion_mp}
\textit{Matrix Profiles} are well suited to detect anomalies in data with periodic characteristics while tolerating a certain amount of noise.
The predominant benefit is the ease of use as there is only one hyperparameter - $m$ - to be defined.
The algorithm is robust to changes in $m$ so that fine-tuning is rarely necessary.
Furthermore,
efficient implementations of the distance calculation allow for in-time calculation of distances and therefore for real-time discovery of attacks.
The choice of a threshold depends on the characteristics of a data set.
 
\section{\textit{SARIMA} approach}
\label{sec:sarima}
In this section we carry out a \ac{sarima} approach for properly modelling and forecasting short-term future values of time series extracted from regular network traffic with periodical characteristics.
Network intrusion can be identified by capturing data points that vary enough from the prediction.

\subsection{Seasonal ARIMA-processes}\label{sec: SARIMA-model}
A stochastic process $\{X_t\}_{t\in \mathbb{Z}}$ is called seasonal ARIMA-process with period $s$ and denoted by $SARIMA(p,d,q)\times(P,D,Q)_s$ if $\{Y_t\}_{t\in \mathbb{Z}}$ with
\begin{equation}\label{eq:SARIMA-d-D}
Y_t=(1-U^{-1})^d(1-U^{-s})^DX_t \quad\text{for } t\in\mathbb{Z}
\end{equation}
is a stationary \ac{arma} process of the form 
\begin{equation}\label{eq:SARIMA-s}
A(U^{-1})F(U^{-s})Y_t=D(U^{-1})G(U^{-s})\epsilon_t, \quad t\in \mathbb{Z},
\end{equation}
where
$\{\epsilon_t \}_{t\in\mathbb{Z}}$
is the innovation process which is supposed to be white noise,
i.e.\ a series of uncorrelated random variables with zero mean and finite variance
$\sigma_\epsilon^2$,
$U$ denotes the shift operator,
i.e., 
 $U:X_t\mapsto X_{t+1}$,
and $A,F,D,G$ refer to the characteristic polynomials related to the $ARMA$-process,
i.e.
\begin{equation}\label{eq:SARIMA-polyn.}
\begin{aligned}
A(z)&=1-\sum_{k=1}^{p}\alpha_k z^k,& D(z)&=1+\sum_{k=1}^{q}\theta_k z^k,
\\
F(z)&=1-\sum_{k=1}^{P}\phi_k z^k,& G(z)&=1+\sum_{k=1}^{Q}\gamma_k z^k.
\end{aligned}
\end{equation}
(\cite[Chap. 9.1.3]{Box.2013})
We assume for simplicity that the innovation process $\{\epsilon_t \}_{t\in\mathbb{Z}}$ is Gaussian white noise, i.e., $\epsilon_t \sim \mathcal{N}(0,\sigma_\epsilon^2)$ for all $t\in \mathbb{Z}$.  

\subsection{Data adaptive model choice}

We consider the three data sets introduced in Section~\ref{sec:ds_for_ad}:
\textit{ds1},
\textit{ds2} and \textit{ds3}.
In order to apply asymptotic results from mathematical statistics,
it is reasonable to use the normal part of the data without intrusion (i.e.,traffic labeled as "0") from the largest dataset (i.e.,with the longest duration) \textit{ds2} as long-term observation to model regular traffic.
This also enables us to highlight the benefit of using the \ac{sarima} approach for noisy data.
The only intrusion in the aforementioned data set is injected during the time interval from the 289th to the 290th second without any effect on the subsequent traffic.
We therefore choose the time series of data points obeserved from the 300th second to the 670th second of that data set for modelling the normal network behavior (training data).
We then apply our forecast model to the remainder of available data (test data) for measuring prediction errors so as to identify attacks. \\ \par
\begin{figure}
	\centering
	
	\begin{subfigure}{.85\linewidth}
		\includegraphics[width=\linewidth]{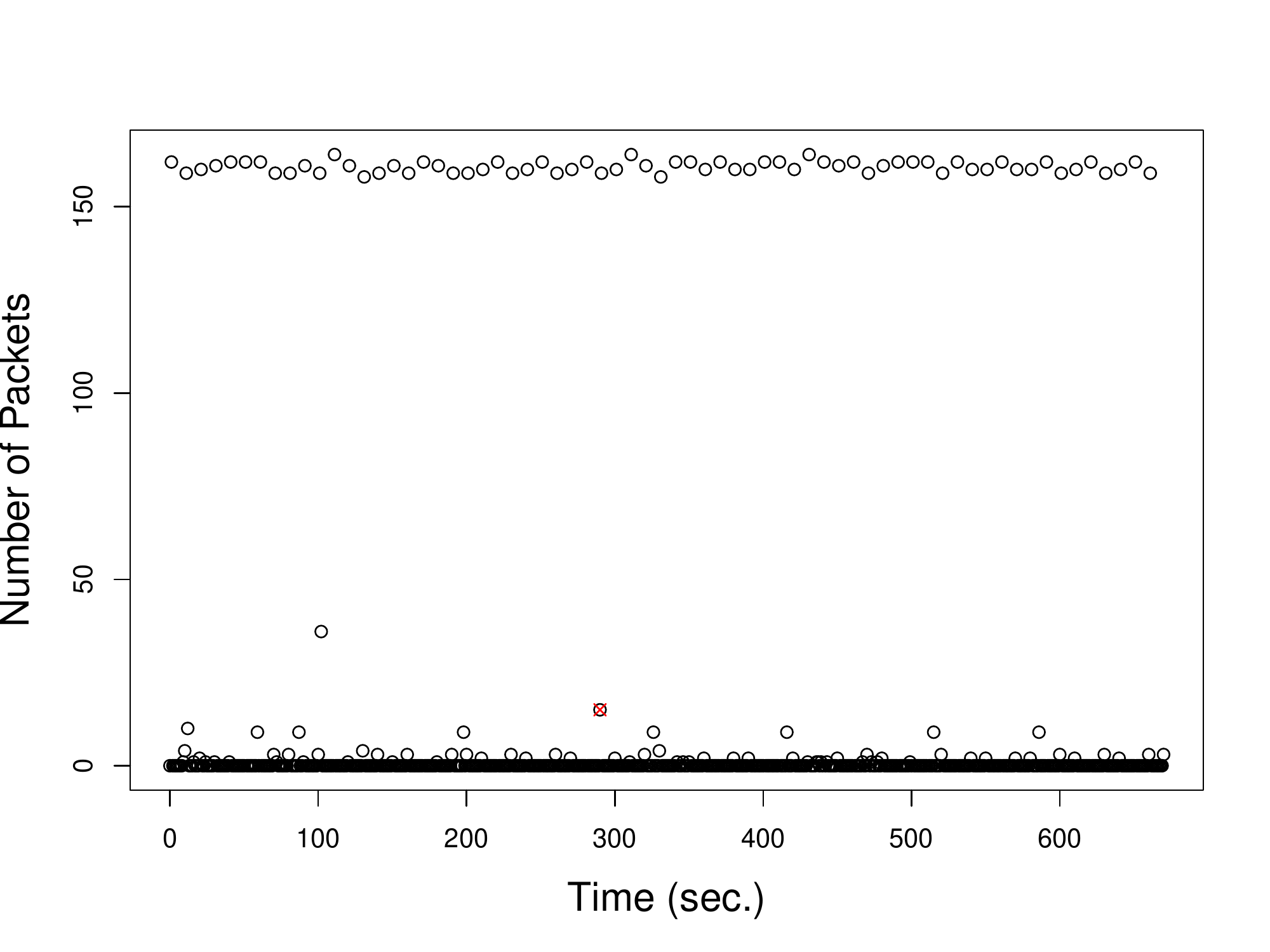}
		\caption{Number of Packets}
	\end{subfigure}
	
	\begin{subfigure}{.85\linewidth}
		\includegraphics[width=\linewidth]{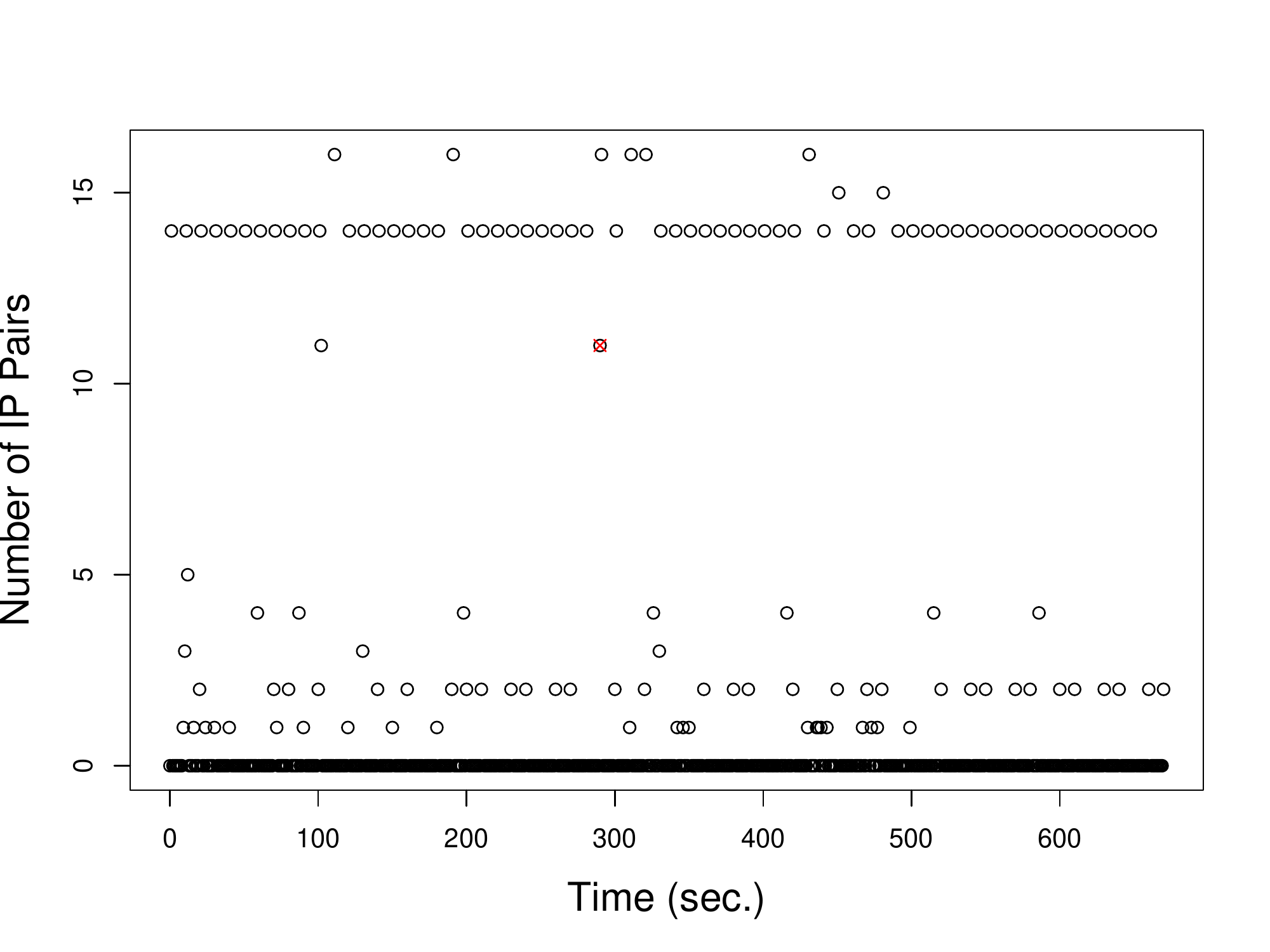}
		\caption{Number of IP Pairs}
	\end{subfigure}
	
	\begin{subfigure}{.85\linewidth}
		\includegraphics[width=\linewidth]{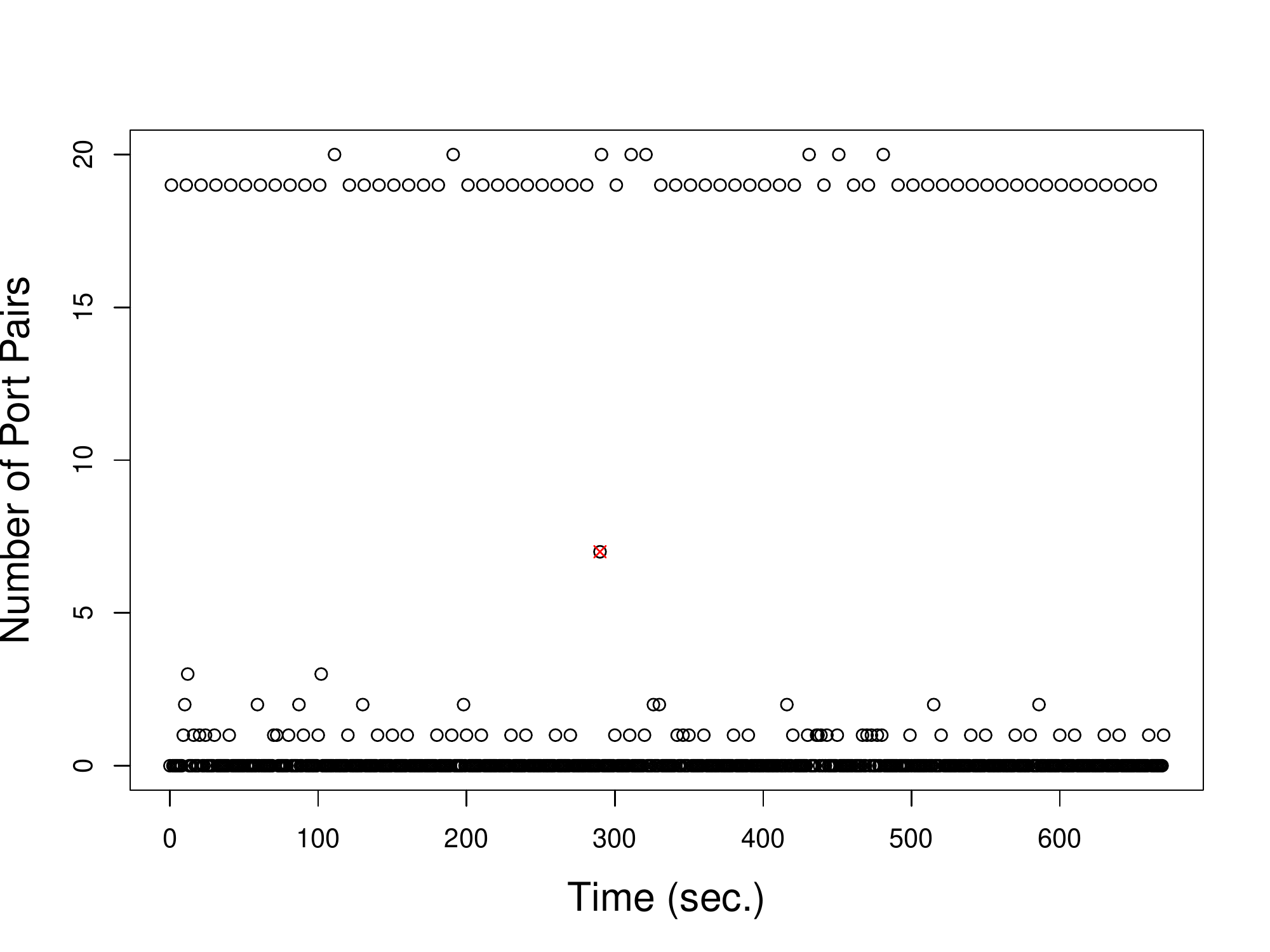}
		\caption{Number of Port Pairs}
	\end{subfigure}
	\caption{Intrusion (crossed out) among other anomalies}
	\label{fig:plot-send}
\end{figure}

In order to bring out the effect of attack traffic and avoid recording too much noise, we capture the increase in the number of sent packets and the number of active IP and port pairs separately at one-second intervals (instead of smaller e.g.\ 10\,ms or 100\,ms intervals). From the large sample correlations between the above crucial features, namely $corr(\text{packet, port})=0.9987$, $corr(\text{port, IP})=0.9922$ and $corr(\text{packet, IP})=0.9878$ (regular traffic from \textit{ds2}), we deduce strong linear dependency of them on one another.  However, since the vast majority of the data is captured during normal system operation, it is not clear how meaningful these numbers are once the actual anomalies occur.  In fact, it turns out that using the number of active port pairs for modeling and testing yields more accurate result on intrusion detection compared to using the other two features.
This is due to the fact that, except for \textit{ds1}, manual operations are included throughout the measurement which can cause considerable fluctuations in the number of packets and IP pairs leading to false positive results during intrusion detection, cf. Figure~\ref{fig:plot-send}-(a), -(b).
In contrast,
the port topology is in general not significantly affected by ordinary manual operations,
cf. Figure~\ref{fig:plot-send}-(c),
which provides us with a feature that is robust against this type of non-intrusion anomalies.
In the sequel,
we only present a detailed treatment of the time series of number of active port pairs per second which is denoted by
$\{X^\prime_t\}_{t=1,...,N}$
with
$N=670-300=370\text{ (sec.)}$
extracted from \textit{ds2} for adjusting our \ac{sarima} model in Section~\ref{sec: SARIMA-model}. 

\begin{figure}
	\centering
	\includegraphics[width=.85\linewidth]{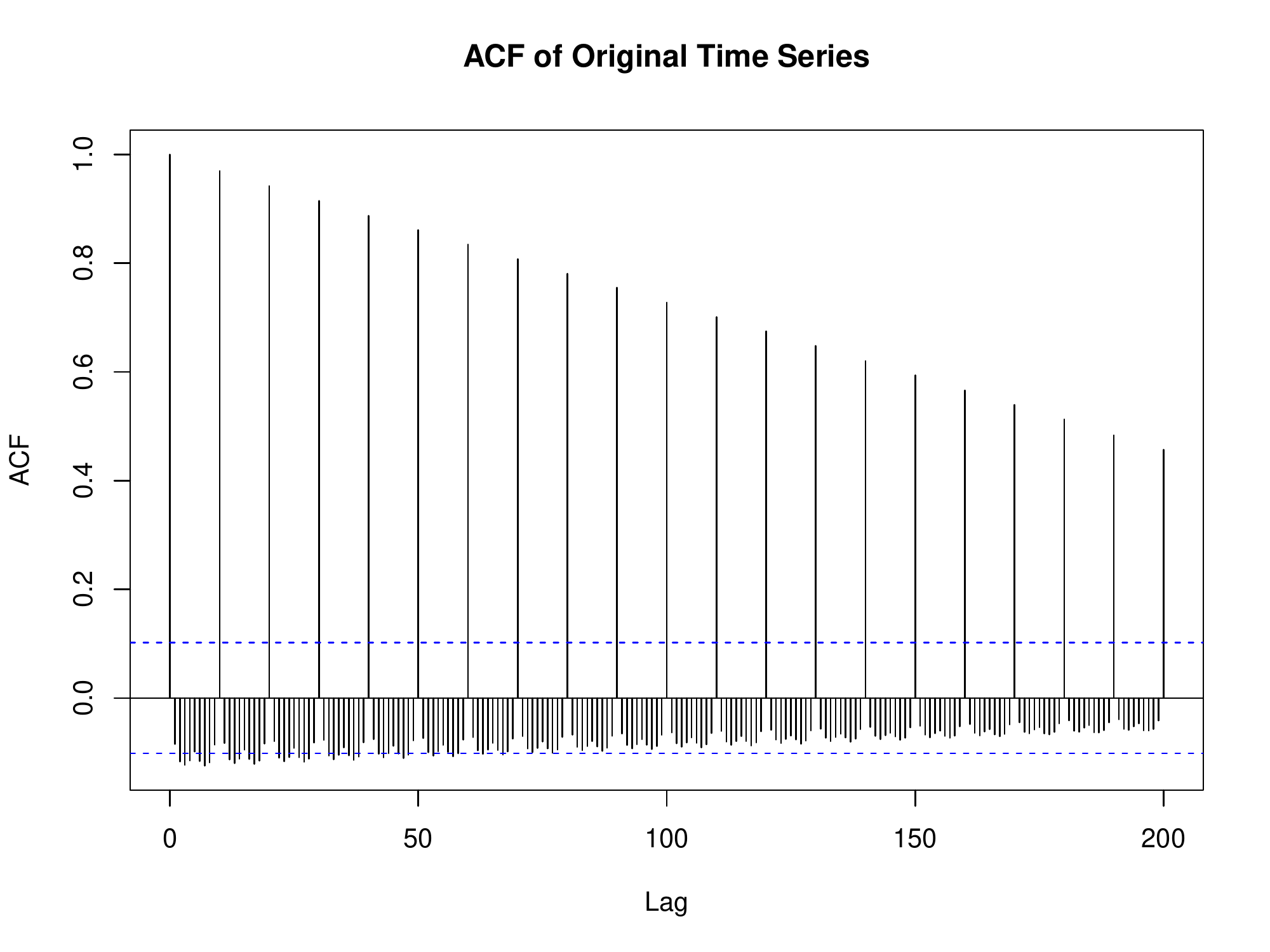}
	\caption{Autocorrelogram of time series $X^\prime$}
	\label{fig:acf}
\end{figure}

First,
aiming to confirm the periodical character induced by the predefined 10-seconds polling interval,
we start our approach by analysing the autocorrelogram of the time series
$\{X^\prime_t\}_{t=1,...,N}$,
which is the plot of sample autocorrelations expressed in terms of
\begin{equation}\label{eq:rho}
\begin{split}
\hat{\rho}^{X^\prime}_{\tau}=\frac{\hat{r}^{X^\prime}_{\tau}}{\hat{r}^{X^\prime}_0}\quad  \text{with} \quad \hat{r}^{X^\prime}_{\tau}=\frac{1}{N}\sum_{k=1}^{N-\tau}(X^\prime_k-\bar{X}^\prime_N)(X^\prime_{\tau+k}-\bar{X}^\prime_N)
\end{split}
\end{equation}
against lags $\tau=0,1,...,N-1$,
cf. Figure~\ref{fig:acf}.
The sample autocorrelation oscillates with constant frequency of ten seconds,
which indicates the existence of a periodical component with $s=10$ in (\ref{eq:SARIMA-s}).
In order to remove the heteroscedasticity of data in respect of state dependent variance and motivated by regression models with time series error terms \cite[Chap. 9.5.1]{Box.2013},
we transform the original time series into a seasonally centered one
$\{X_t\}_{t=1,\ldots,N}$,
that is,  
\begin{equation}\label{eq:transf-seasonal}
X_{j+us}:=X^\prime_{j+us}-\frac{1}{N/s}\sum_{i=0}^{N/s-1}X^\prime_{j+is},
\end{equation}
for $j=1,\ldots,s-1$, $u=0,\ldots,N/s-1$.
Since our time series is not expected to have any time dependent trend, we choose $d = D = 0$ in (\ref{eq:SARIMA-d-D}),
thereby obtaining a $SARIMA(p, 0, q) \times (P, 0, Q)_{10}$ model with the time series
$\{ Y_{t} \}_{t = 1, \ldots, N}$
given by
$Y_{t} := X_{t}$
for all
$t = 1, \ldots, N$.
(Note that $\bar{Y}_{N} = \bar{X}_{N} = 0$.)

In order to determine the orders of the moving average part $(Q,q)$ and the autoregressive part $(P,p)$,
we analyse the autocorrelogram and partial autocorrelogram related to
$\{Y_t\}_{t=1,\ldots,N}$,
respectively.
The sample autocorrelations $\hat{\rho}^Y_\tau$ for $\tau=0,1,...,N-1$ can be computed as in (\ref{eq:rho}),
whereas the sample partial correlations $\hat{\pi}^Y_{\tau}$ can be determined by means of the Yule-Walker equation~\cite[Chap. 3.2.6]{Box.2013},
which gives
\begin{gather*}
\hat{\pi}^Y_{\tau}=(\hat{R}^Y_{\tau})^{-1}\hat{r}^Y(\tau) \\ \text{with} \quad \hat{R}^Y_{\tau}=(\hat{r}^Y_{t-k})_{1\leq t,k\leq \tau},\quad \hat{r}^Y(\tau)=(\hat{r}^Y_1,...\hat{r}^Y_{\tau})^T. 
\end{gather*}
We then obtain possible choices of $P$ and $Q$ by considering $\hat{\pi}^Y_{ks}$ and $\hat{\rho}^Y_{ks}$ for $k\geq 0$ and picking those values of $k$ where $\hat{\pi}^Y_{(k+1)s}$ and $\hat{\rho}^Y_{(k+1)s}$ begin to fall exponentially towards zero, respectively. Moreover, analysing $\hat{\pi}^Y_1,...,\hat{\pi}^Y_{s-1}$ and $\hat{\rho}^Y_1,...,\hat{\rho}^Y_{s-1}$ in the above manner provides possible choices of $p$ and $q$, respectively. The choice of hyperparameters $P,Q,p,q$ is not unique in general; a final decision among the possible candidates can be made, for instance, by means of the Akaike information criterion (AIC). Since our focus is not on delivering the best possible forecast but on detecting outliers and as these can also occur in early periods (cf. e.g. \textit{ds1}), also doing so with the least possible amount of preceding data, we make a compromise of retaining accuracy and computational simplicity and accept the combination of orders $(p=4,q=0)\times(P=1,Q=0)$. Summing up, the above consideration leads to our choice of the model $SARIMA(4,0,0)\times(1,0,0)_{10}$ for $\{Y_t\}_{t=1,\ldots,N}$.

\subsection{Parameter estimation}\label{sec:LS}

Since we have assumed the innovation $\{\epsilon_t\}_{t}$ to be Gaussian white noise,
we use the least squares estimation which approximately delivers the asymptotically efficient maximum likelihood estimate of the parameters
$\alpha:=(\alpha_1,\ldots,\alpha_p)$,
$\phi:=(\phi_1,\ldots,\phi_P)$ and $\sigma_\epsilon$ for $SARIMA(p=4,0,0)\times(P=1,0,0)_{10}$.
The functional we are going to minimize reads as
\[
f(\alpha,\phi):= \sum_{t=Ps+p+1}^{N}\epsilon_t(\alpha,\phi)^2
\]
with
\begin{multline}\label{eq:pred}
Y^*_t(\alpha,\phi)=\\ \sum_{j=1}^{p}\alpha_j Y_{t-j}+\sum_{k=1}^{P}\phi_k Y_{t-sk}-\sum_{j=1}^{p}\sum_{k=1}^{P}\alpha_j\phi_kY_{t-sk-j},
\end{multline}
\begin{equation}\label{eq:residual}
\epsilon_t(\alpha,\phi)_t=Y_t-Y^*_t(\alpha,\phi), \quad t>Ps+p+1.
\end{equation}
It holds for the least squares estimate that
\begin{gather*}
(\hat{\alpha},\hat{\phi})=\argmin_{\alpha\in\mathbb{R}^p,\phi \in\mathbb{R}^P}f(\alpha,\phi),\\ \hat{\sigma}_\epsilon^2=MSE=\frac{f(\hat{\alpha},\hat{\phi})}{N-Ps-p}
\end{gather*}
with $p=4$, $P=1$, $s=10$.
We solve the minimization problem numerically by means of a gradient descent procedure.
The result of the above approximation is presented in Table~\ref{tab:LS}.
We then obtain the one-step-ahead prediction of $\{Y_t\}_{t}$ and the residuals
$\{\epsilon_t(\hat{\alpha},\hat{\phi})\}_{t}$
in terms of (\ref{eq:pred}, \ref{eq:residual}).

\begin{table}[h!]
\renewcommand{\arraystretch}{1.3}
\caption{LS Estimates}
\label{tab:LS}
\centering
\begin{tabular}{l r}
\toprule
			\textbf{Parameter} & \textbf{Estimated Value} \\
			$\hat{\alpha}_1$ & $        -1.0997\,\mathrm{e}\,{-2}$\\
			$\hat{\alpha}_2$ & $        -9.9894\,\mathrm{e}\,{-4}$\\
			$\hat{\alpha}_3$ & $\phantom-6.8105\,\mathrm{e}\,{-4}$\\
			$\hat{\alpha}_4$ & $\phantom-1.3458\,\mathrm{e}\,{-1}$\\
			$\hat{\phi}_1$ & $-1.1170\,\mathrm{e}\,{-1}$\\
$\hat{\sigma}_{\epsilon}^2$ & $\phantom-1.0239\,\mathrm{e}\,{-1}$ \\
\bottomrule
\end{tabular}
\end{table}

In a final step, let us verify that indeed our model is reasonable in the sense that the residuals $\epsilon_t(\hat{\alpha},\hat{\phi})$,
$t=Ps+p+1,\ldots,N$ are white noise.
To this end, we consider their sample autocorrelations
$\hat{\rho}^\epsilon_1, \hat{\rho}^\epsilon_2, \ldots$
and conduct the \textit{Ljung-Box} test~\cite[Chap. 8.2.2]{Box.2013} that uses as test statistic
\[
Q:=N(N+2)\sum_{k=1}^{H}\frac{(\hat{\rho}_{k}^{\epsilon})^2}{N-k}
\]
where we choose $H=\lfloor 2\sqrt{N}\rfloor$ and critical region $\{Q > q_{1-\alpha}(\chi^2_{H-(Ps+p)}) \}$ for a significance level of $\alpha=0.05$.  It turns out that in our setting, $Q=33.40707$, $q_{1-\alpha}(\chi^2_{H-(Ps+p)})=36.9982$.

\subsection{Intrusion detection}

Having modelled the regular traffic, we now detect network intrusion as follows.
For each test data set,
we first transform the relevant time series
$\{Z^\prime_t\}_t$
into a seasonally centred one $\{Z_t\}_t$ in the sense of (\ref{eq:transf-seasonal}) with fixed sample seasonal means obtained from the \emph{training} data $\{X^\prime_t\}_t$.
Then we apply the $SARIMA(4,0,0)\times(1,0,0)_{10}$ model and the estimated coefficients $\hat{\alpha},\hat{\phi}$ from Section~\ref{sec:LS} to the transformed test time series $\{Z_t\}_t$ and compute the one-step prediction errors $e^Z_t$ in terms of (\ref{eq:pred}, \ref{eq:residual}) for $t=Ps+p+1, Ps+p+2,\ldots$.
Due to the different nature of the three test data sets in respect of manual operations,
which belong to non-intrusion anomalies,
we set individual thresholds for evaluating prediction errors in different data sets.
For instance,
since no manual operations are conducted in \textit{ds1},
we can choose the $0.9995$-quantile of $\mathcal{N}(0,\hat{\sigma}^2_\epsilon)$ ($=1.05293$) as threshold (corresponding to the confidence interval for a significance level of $0.001$) for the absolute value of the prediction errors related to the number of port pairs in that data set.
In contrast, more manual operations are observed in \textit{ds2} so that we set three times the $0.9995$-quantile of $\mathcal{N}(0,\hat{\sigma}^2_\epsilon)$ as threshold instead.
As soon as the absolute prediction error $\abs{e_{t_0}^Z}$ exceeds the threshold at some time $t_0$,
the traffic related to that state $Z_{t_0}$ in the time series is classified as an intrusion.
In order to prevent consequent errors, we also remove the detected attack traffic from the data immediately after detection,
i.e.,
we replace the detected outlier $Z_{t_0}$ by the corresponding regular value before we continue the detection procedure for $t>t_0$. 

The final results of the above approach are summarised in Tables~\ref{tab:res-Send}, \ref{tab:res-Mov}, \ref{tab:res-CnC}. It turns out that, by chosing the proper threshold for each data set, all attack traffic can be accurately detected within one second while producing only a single false positive.

\begin{table}[h!]
\renewcommand{\arraystretch}{1.3}
\caption{Result on \textit{ds2}, Threshold for $\abs{e_t^Z}$: $3q_{0.9995}(\mathcal{N}(0,\hat{\sigma}_\epsilon^2)) $}
\label{tab:res-Send}
\centering
\begin{tabular}{r r}
\toprule
			\textbf{Begin of Attack Traffic (s)} & \textbf{First Detection Time (s)} \\
			$289.4079$ & $290$ \\
\bottomrule
\end{tabular}
\end{table}

\begin{table}[h!]
\renewcommand{\arraystretch}{1.3}
\caption{Result on \textit{ds1}, Threshold for $\abs{e_t^Z}$: $q_{0.9995}(\mathcal{N}(0,\hat{\sigma}_\epsilon^2)) $}
\label{tab:res-Mov}
\centering
\begin{tabular}{r r}
\toprule
			\textbf{Begin of Attack Traffic (s)} & \textbf{First Detection Time (s)} \\
			$10.8980$ & N/A\footnotemark\\
			$32.9679$ & $34$\\
			$71.5955$ & $72$\\
			$93.6086$ & $94$ \\
\bottomrule
\end{tabular}
\end{table}

\begin{table}[h!]
\renewcommand{\arraystretch}{1.3}
\caption{Result on \textit{ds3}, Threshold for $\abs{e_t^Z}$: $q_{0.9995}(\mathcal{N}(0,\hat{\sigma}_\epsilon^2)) $}
\label{tab:res-CnC}
\centering
\begin{tabular}{r r}
\toprule
			\textbf{Begin of Attack Traffic (s)} & \textbf{First Detection Time (s)} \\
			$44.3293$ & $45$ \\
			- & 63\\
			$64.1758$ & $65$ \\		
\bottomrule
\end{tabular}
\end{table}
\footnotetext{As this anomaly occurres before $Ps + p$ seconds, the model is not applicable here.}

\subsection{Discussion}
Overall,
the main advantage of the \ac{sarima} approach is that it is still powerful in the presence of noise.
It also provides a parsimonious presentation of time series with periodical characteristics,
using few parameters.
However,
it requires individual model adjustment each time dealing with a new time series.

\section{Long Short Term Memory}
\label{sec:lstm}
\ac{lstm} is a kind of neural network proposed by \textit{Hochreiter and Schmidhuber} in 1997 to overcome the vanishing gradient problem~\cite{Hochreiter.1997}.
This problem occurs when long-term dependencies are not considered accordingly by a recurrent neural network.
The network ``forgets" the events and cannot correlate dependencies.
The concept of \ac{lstm} is presented in Subsection~\ref{ssec:introduction_lstm}.
After that,
the evaluation of \ac{lstm} on the data sets is provided in Subsection~\ref{ssec:eval_lstm},
followed by a discussion of advantages and disadvantages in Subsection~\ref{ssec:discussion_lstm}.

\subsection{Introduction to LSTM}
\label{ssec:introduction_lstm}
\begin{figure}[!ht]
\centering
\includegraphics[width=0.5\textwidth]{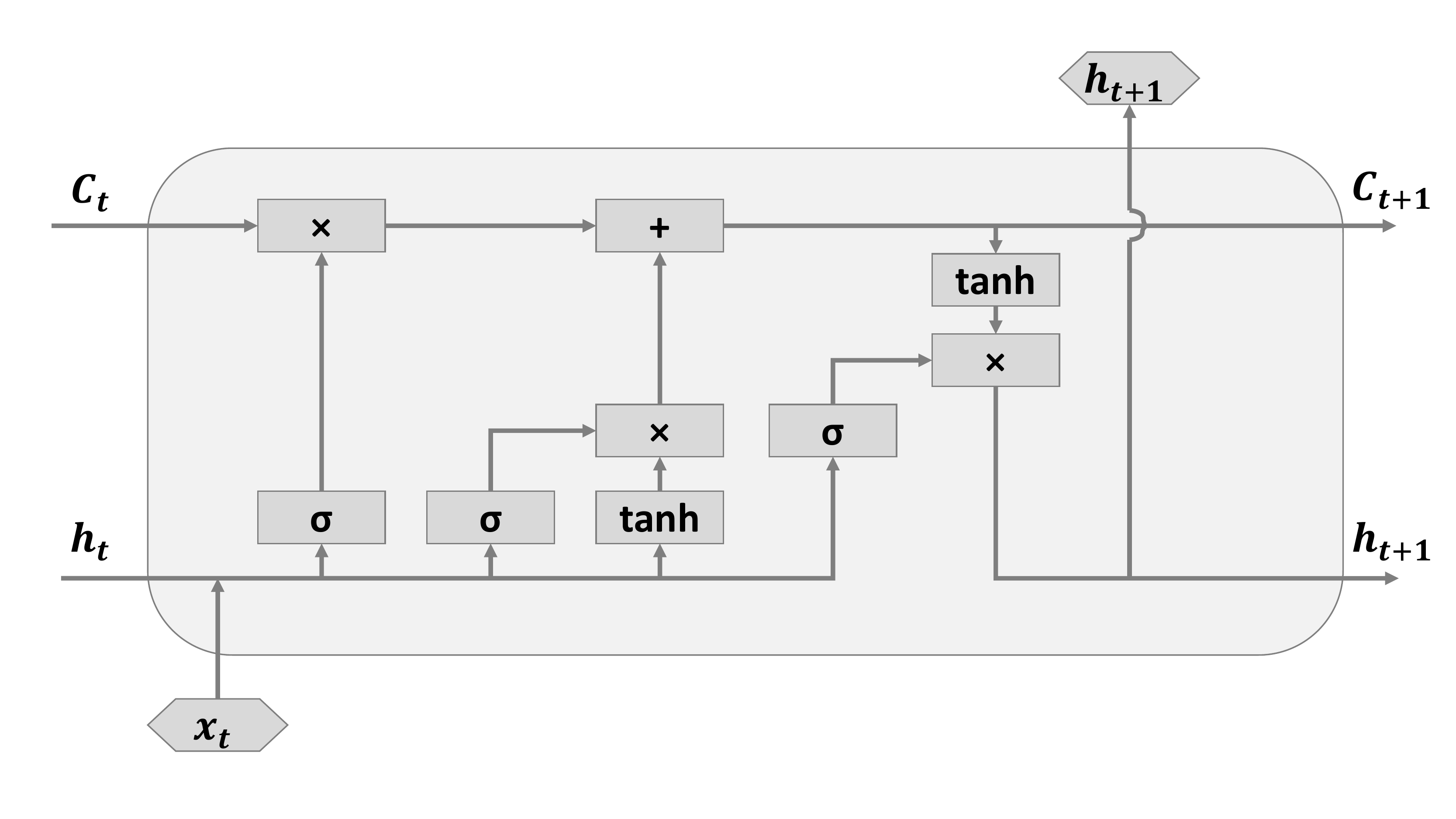}
\caption{Schematic of an LSTM Cell}
\label{fig:lstm_cell}
\end{figure}

\ac{lstm} is a kind of neural network designed to keep information over long periods of time.
Due to this ability,
\ac{lstm} networks need the ability to reset parameters.
This can be done with forget gates~\cite{Gers.1999}.
\ac{lstm} networks consist of cells that are interconnected.
One such cell is depicted in Figure~\ref{fig:lstm_cell}.
The representation of \textit{Olah} is followed in this work~\cite{Olah.2015}. \par
$x_{t}$ is the input,
$h_{t}$ the output and $C_{t}$ the cell state at time step $t$.
They are passed as inputs to the adjacent \ac{lstm} cell.
$W$ is a weight vector and $b$ is a bias.
Furthermore,
there are the functions $i_{t}$,
$f_{t}$ and $o_{t}$.
They represent the update function,
the forget function and the output function.
To determine the values,
activation functions are used.
They are either the \textit{tangens hyperbolic (tanh)} or the \textit{sigmoid function ($\sigma$)}.
The forget value is calculated according to (\ref{eq:forget_gate}).
\begin{equation}\label{eq:forget_gate}
\begin{split}
f_{t} = \sigma(W_{f} \cdot [h_{t-1},x_{t}] + b_{f})
\end{split}
\end{equation}
$h_{t}$ is derived from $h_{t-1}$ and $x_{t-1}$,
as are $i_{t}$ and $\tilde{C}_{t}$.
They are calculated according to (\ref{eq:update_func}) and (\ref{eq:update_state}).
\begin{equation}\label{eq:update_func}
\begin{split}
i_{t} = \sigma(W_{i} \cdot [h_{t-1}, x_{t}] + b_{i})
\end{split}
\end{equation}
\begin{equation}\label{eq:update_state}
\begin{split}
\tilde{C}_{t} = tanh(W_{C} \cdot [h_{t-1}, x_{t}] + b_{C})
\end{split}
\end{equation}
$\tilde{C}_{t}$ is an internal state used to calculate the cell state,
as shown in (\ref{eq:cell_state}).
\begin{equation}\label{eq:cell_state}
\begin{split}
C_{t} = f_{t} \ast C_{t-1} + i_{t} \ast \tilde{C}_{t}
\end{split}
\end{equation}
The output variable $o_{t}$ and the output gate $h_{t}$ are calculated according to (\ref{eq:output_value}) and (\ref{eq:output_gate}).
\begin{equation}\label{eq:output_value}
\begin{split}
o_{t} = \sigma(W_{o} \cdot [h_{t-1}, x_{t}] + b_{o})
\end{split}
\end{equation}
\begin{equation}\label{eq:output_gate}
\begin{split}
h_{t} = o_{t} \ast tanh(C_{t})
\end{split}
\end{equation}
Four different parameterisations have been employed in this work:
The length of input sequences has been set to 10 and 20,
the number of layers has been set to 1 and 3.
Furthermore,
400 neurons have been used.
The training has been performed on the \textit{ds1} that has been stripped of malicious events using \numprint{20000} iterations with a learning rate of \numprint{0.001} and batch sizes of 50.
The \ac{lstm} networks have been used as predictors.
They predicted the next event which was then compared to the actual event.
In order to prevent consequent errors,
the detected attack traffic was removed from the data immediately after detection before continuing the prediction process.
All prediction errors were calculated as the absolute difference of predicted and actual value.
After that,
two threshold values were calculated.
The first threshold was the minimal value,
so that all malicious evens are above the threshold (\textit{MA}).
The second threshold was chosen so that all non-malicious events were below the threshold value (\textit{NM}).
Only the best performing set ups are presented in the following.

\subsection{Evaluation of LSTM}
\label{ssec:eval_lstm}
After setting the threshold as discussed in the previous subsection,
the \textit{true positives} ($t_{p}$),
\textit{false positives} ($f_{p}$),
\textit{true negatives} ($t_{n}$) and \textit{false negatives} ($f_{n}$) are calculated.
They are used to calculate the F1-score (\ref{eq:f-measure}) with precision (\ref{eq:precision}) and recall (\ref{eq:recall}),
as well as the accuracy (\ref{eq:accuracy}).
These metrics were evaluated for each feature,
packet count (\textit{PC}),
IP pairs (\textit{IP}) and port pairs (\textit{PP}) respectively.
The performance of \ac{lstm} networks is presented in Table~\ref{tab:perf_lstm}.
\begin{equation}
\label{eq:f-measure}
F_{1} = 2 \cdot \dfrac{precision \cdot recall}{precision + recall}
\end{equation}
\begin{equation}
\label{eq:precision}
precision = \dfrac{t_{p}}{t_{p}+f_{p}}
\end{equation}
\begin{equation}
\label{eq:recall}
recall = \dfrac{t_{p}}{t_{p}+f_{n}}
\end{equation}
\begin{equation}
\label{eq:accuracy}
accuracy = \dfrac{t_{p} + t_{n}}{t_{p}+f_{p} + t_{n}+f_{n}}
\end{equation}
\begin{table*}[h!]
\renewcommand{\arraystretch}{1.3}
\caption{Performance of \ac{lstm}}
\label{tab:perf_lstm}
\centering
\begin{tabular}{l c c c c c c c c c c}
\toprule
\multicolumn{2}{c}{\multirow{ 2}{*}{\textbf{Feature}}} & \phantom{a} & \multicolumn{2}{c}{\textbf{\textit{ds1}}} & \phantom{a} & \multicolumn{2}{c}{\textbf{\textit{ds2}} } & \phantom{a} & \multicolumn{2}{c}{\textbf{\textit{ds3}}}\\
 & & & \textit{MA (\%)} & \textit{NM (\%)} & & \textit{MA (\%)} & \textit{NM (\%)} & & \textit{MA (\%)} & \textit{NM (\%)} \\
 \cmidrule{1-2} \cmidrule{4-5} \cmidrule{7-8} \cmidrule{10-11}
 \multirow{2}{*}{\textit{PC}} & \textit{Accuracy} & & 98.4293 & 97.9058 & & 99.5529 & 99.8510 & & 95.7746 & 95.7746 \\
 &  \textit{F1-Score} & & 86.9565 & 75.0000 & & 40.0000 & 0 & & 80.0000 & 66.6667 \\
 \multirow{2}{*}{\textit{IP}} & \textit{Accuracy} & & 90.0524 & 98.9529 & & 99.9529 & 99.8510 & & 98.5915 & 92.3077 \\ 
 &  \textit{F1-Score} & & 51.2821 & 88.8889 & & 66.6667 & 0 & & 92.3077 & 28.5714 \\
 \multirow{2}{*}{\textit{PP}} & \textit{Accuracy} & & 98.9529 & 99.4764 & & 99.1058 & 99.8510 & & 98.5915 & 92.9577 \\
 &  \textit{F1-Score} & & 90.9091 & 94.7368 & & 25.0000 & 0 & & 92.3077 & 28.5714 \\
\bottomrule
\end{tabular}
\end{table*}

\subsection{Discussion}
\label{ssec:discussion_lstm}
Table~\ref{tab:perf_lstm} shows that the accuracy is always higher than the F1-score.
This is due to the fact that $t_{p}$ as well as $t_{n}$ are considered in calculating the accuracy,
while the F1-score puts an emphasis on $t_{p}$.
As shown in Table~\ref{tab:data_sets},
the malicious events are magnitudes smaller than the non-malicious ones.
This is a common problem in anomaly detection,
as an anomaly happens less frequently~\cite{Chandola.2009}.
It leads to a negligible $t_{p}$ on the accuracy,
most prevalently shown with \textit{ds2}:
The only attack was considered as non-malicious and thus a $f_{n}$,
but the accuracy according to (\ref{eq:accuracy}) appears to be almost perfect. 
Nevertheless, 
\ac{lstm} performed relatively good analysing \textit{ds1} and \textit{ds3} considering some features.

\section{Conclusion}
\label{sec:conclusion}
In this work,
three algorithms for time series-based anomaly detection were evaluated on a synthetic data set containing network traffic of an industrial use case.
They were analysed with respect to their capability in detecting attacks that were introduced to that data set,
as well as the effort required to parameterise and train the algorithms. \\ \par
The \textit{Matrix Profile} approach performs very well in comparison.
Most attacks can be found with few false positives. 
Furthermore,
the parametrisation effort requires only one hyperparameter that is robust to change.
Additionally,
the training data set is small.
There are some noteworthy extensions to \textit{Matrix Profiles},
e.g. multidimensionality~\cite{Yeh.2017} and concepts for the integration of domain knowledge~\cite{Dau.2017},
providing further applications to intrusion detection. \\ \par
The \ac{sarima} approach provides high forecasting accuracy and detection performance with few parameters.
The theory of time series analysis from mathematical statistics provides a clearly defined procedure for the data adaptive model choice and the model adequacy check,
e.g. the \textit{Ljung-Box} test. \\ \par
Despite the high computational effort for training a large number of parameters,
the performance of \ac{lstm} network is less convincing than that of the other two algorithms.
It strongly depends on the nature of the data.
Over- or underfitting are important issues that have to be addressed.
Furthermore,
\ac{lstm} networks require the specification of many hyperparameters,
e.g. number of layers,
choice of activation function and learning rate,
whose fine tuning is tedious and often depends on intuition and experience.
That makes them unsuited for non-experts. \\ \par
In summary,
the time series-based anomaly detection methods discussed in this work are effective in detecting cyber attacks in industrial network traffic.
Their efficiency could be improved by incorporating context information~\cite{Duque_Anton.2017c} such as authentication when operating manually,
and sensible aggregation of this information~\cite{Duque_Anton.2017b}.

\section*{Acknowledgments}
This work has been supported by the Federal Ministry of Education and Research of the Federal Republic of Germany (Foerderkennzeichen KIS4ITS0001, IUNO).
The authors alone are responsible for the content of the paper.



\begin{thebibliography}{10}
\providecommand{\url}[1]{#1}
\csname url@samestyle\endcsname
\providecommand{\newblock}{\relax}
\providecommand{\bibinfo}[2]{#2}
\providecommand{\BIBentrySTDinterwordspacing}{\spaceskip=0pt\relax}
\providecommand{\BIBentryALTinterwordstretchfactor}{4}
\providecommand{\BIBentryALTinterwordspacing}{\spaceskip=\fontdimen2\font plus
\BIBentryALTinterwordstretchfactor\fontdimen3\font minus
  \fontdimen4\font\relax}
\providecommand{\BIBforeignlanguage}[2]{{%
\expandafter\ifx\csname l@#1\endcsname\relax
\typeout{** WARNING: IEEEtran.bst: No hyphenation pattern has been}%
\typeout{** loaded for the language `#1'. Using the pattern for}%
\typeout{** the default language instead.}%
\else
\language=\csname l@#1\endcsname
\fi
#2}}
\providecommand{\BIBdecl}{\relax}
\BIBdecl

\bibitem{Duque_Anton.2017a}
S.~Duque~Anton, D.~Fraunholz, C.~Lipps, F.~Pohl, M.~Zimmermann, and H.~D.
  Schotten, ``Two decades of scada exploitation: A brief history,'' in
  \emph{2017 IEEE Conference on Application, Information and Network Security
  (AINS)}, November 2017, pp. 98--104.

\bibitem{Igure.2006}
V.~M. Igure, S.~A. Laughter, and R.~D. Williams, ``{Security issues in SCADA
  networks},'' \emph{Computers {\&} Security}, vol.~25, pp. 498--506, 2006.

\bibitem{Zhu.2011}
\BIBentryALTinterwordspacing
B.~Zhu, A.~Joseph, and S.~Sastry, ``A taxonomy of cyber attacks on scada
  systems,'' in \emph{Proceedings of the 2011 International Conference on
  Internet of Things and 4th International Conference on Cyber, Physical and
  Social Computing}, ser. ITHINGSCPSCOM.\hskip 1em plus 0.5em minus 0.4em\relax
  Washington, DC, USA: IEEE Computer Society, 2011, pp. 380--388. [Online].
  Available: \url{http://dx.doi.org/10.1109/iThings/CPSCom.2011.34}
\BIBentrySTDinterwordspacing

\bibitem{Mitchell.2014}
\BIBentryALTinterwordspacing
R.~Mitchell and I.-R. Chen, ``A survey of intrusion detection techniques for
  cyber-physical systems,'' \emph{ACM Computing Surveys}, vol.~46, no.~4, pp.
  1--29, Mar. 2014. [Online]. Available:
  \url{http://doi.acm.org/10.1145/2542049}
\BIBentrySTDinterwordspacing

\bibitem{Gupta.2014}
M.~Gupta, J.~Gao, C.~C. Aggarwal, and J.~Han, ``Outlier detection for temporal
  data: A survey,'' \emph{IEEE Transactions on Knowledge and Data Engineering},
  vol.~26, no.~9, pp. 2250--2267, September 2014.

\bibitem{Sperotto.2010}
A.~Sperotto, G.~Schaffrath, R.~Sadre, C.~Morariu, A.~Pras, and B.~Stiller, ``An
  overview of ip flow-based intrusion detection,'' \emph{IEEE Communications
  Surveys Tutorials}, vol.~12, no.~3, pp. 343--356, March 2010.

\bibitem{Lu.2009}
W.~Lu and A.~A. Ghorbani, ``Network anomaly detection based on wavelet
  analysis,'' \emph{EURASIP J. Adv. Signal Process}, vol. 2009, pp. 1--16,
  January 2009.

\bibitem{Barford.2002}
P.~Barford, J.~Kline, D.~Plonka, and A.~Ron, ``A signal analysis of network
  traffic anomalies,'' in \emph{Proceedings of the 2Nd ACM SIGCOMM Workshop on
  Internet Measurment}, ser. IMW '02.\hskip 1em plus 0.5em minus 0.4em\relax
  New York, NY, USA: ACM, 2002, pp. 71--82.

\bibitem{Celenk.2008}
M.~Celenk, T.~Conley, J.~Graham, and J.~Willis, ``Anomaly prediction in network
  traffic using adaptive wiener filtering and arma modeling,'' in \emph{2008
  IEEE International Conference on Systems, Man and Cybernetics}, October 2008,
  pp. 3548--3553.

\bibitem{Muenz.2007}
G.~M\"{u}nz, S.~Li, and G.~Carle, ``Traffic anomaly detection using kmeans
  clustering,'' in \emph{In GI/ITG Workshop MMBnet}, 2007.

\bibitem{Kim.2004}
M.-S. Kim, H.-J. Kong, S.-C. Hong, S.-H. Chung, and J.~W. Hong, ``A flow-based
  method for abnormal network traffic detection,'' in \emph{2004 IEEE/IFIP
  Network Operations and Management Symposium}, vol.~1, April 2004, pp.
  599--612.

\bibitem{Sperotto.2008}
A.~Sperotto, R.~Sadre, and A.~Pras, ``Anomaly characterization in flow-based
  traffic time series,'' in \emph{Proceedings of the 8th IEEE International
  Workshop on IP Operations and Management}, ser. IPOM '08.\hskip 1em plus
  0.5em minus 0.4em\relax Berlin, Heidelberg: Springer-Verlag, 2008, pp.
  15--27.

\bibitem{Filonov.2016}
\BIBentryALTinterwordspacing
P.~Filonov, A.~Lavrentyev, and A.~Vorontsov, ``Multivariate industrial time
  series with cyber-attack simulation: Fault detection using an lstm-based
  predictive data model,'' \emph{CoRR}, 2016. [Online]. Available:
  \url{http://arxiv.org/abs/1612.06676}
\BIBentrySTDinterwordspacing

\bibitem{Lin.2017}
\BIBentryALTinterwordspacing
C.-Y. Lin, S.~Nadjim-Tehrani, and M.~Asplund, ``Timing-based anomaly detection
  in {SCADA} networks,'' 2017. [Online]. Available:
  \url{https://pdfs.semanticscholar.org/bc74/ca2e548c1567f4bd8794d480b83a91115f32.pdf}
\BIBentrySTDinterwordspacing

\bibitem{Linda.2009}
O.~Linda, T.~Vollmer, and M.~Manic, ``Neural network based intrusion detection
  system for critical infrastructures,'' in \emph{2009 International Joint
  Conference on Neural Networks}, June 2009, pp. 1827--1834.

\bibitem{Goldenberg.2013}
N.~Goldenberg and A.~Wool, ``Accurate modeling of {Modbus/TCP} for intrusion
  detection in scada systems,'' \emph{International Journal of Critical
  Infrastructure Protection}, vol.~6, no.~2, pp. 63 -- 75, 2013.

\bibitem{Fovino.2010}
I.~N. Fovino, A.~Carcano, T.~De~Lacheze~Murel, A.~Trombetta, and M.~Masera,
  ``{Modbus/DNP3} state-based intrusion detection system,'' in \emph{24th IEEE
  International Conference on Advanced Information Networking and
  Applications(AINA)}, April 2010, pp. 729--736.

\bibitem{Carcano.2009}
A.~Carcano, I.~N. Fovino, M.~Masera, and A.~Trombetta, ``State-based network
  intrusion detection systems for {SCADA} protocols: A proof of concept,'' in
  \emph{CRITIS 2009: Critical Information Infrastructures Security}, vol.
  6027.\hskip 1em plus 0.5em minus 0.4em\relax Springer, Berlin, Heidelberg,
  June 2009, pp. 138--150.

\bibitem{Yang.2014}
Y.~Yang, K.~McLaughlin, S.~Sezer, T.~Littler, E.~G. Im, B.~Pranggono, and H.~F.
  Wang, ``Multiattribute scada-specific intrusion detection system for power
  networks,'' \emph{IEEE Transactions on Power Delivery}, vol.~29, no.~3, pp.
  1092--1102, June 2014.

\bibitem{Lin.2013}
H.~Lin, A.~Slagell, C.~Di~Martino, Z.~Kalbarczyk, and R.~K. Iyer, ``Adapting
  bro into scada: Building a specification-based intrusion detection system for
  the dnp3 protocol,'' in \emph{Proceedings of the Eighth Annual Cyber Security
  and Information Intelligence Research Workshop}, ser. CSIIRW '13.\hskip 1em
  plus 0.5em minus 0.4em\relax New York, NY, USA: ACM, 2013, pp. 5:1--5:4.

\bibitem{Verba.2008}
J.~Verba and M.~Milvich, ``Idaho national laboratory supervisory control and
  data acquisition intrusion detection system (scada ids),'' in \emph{2008 IEEE
  Conference on Technologies for Homeland Security}, May 2008, pp. 469--473.

\bibitem{Oman.2008}
P.~Oman and M.~Phillips, ``Intrusion detection and event monitoring in scada
  networks,'' in \emph{Critical Infrastructure Protection}.\hskip 1em plus
  0.5em minus 0.4em\relax Boston, MA: Springer US, 2008, pp. 161--173.

\bibitem{Staudemeyer.2015}
R.~C. Staudemeyer, ``An overview of ip flow-based intrusion detection,''
  \emph{South African Computer Journal}, no.~56, pp. 136--154, July 2015.

\bibitem{Bontemps.2016}
L.~Bontemps, V.~L. Cao, J.~McDermott, and N.-A. Le-Khac, ``Collective anomaly
  detection based on long short-term memory recurrent neural networks,'' in
  \emph{Future Data and Security Engineering}, vol. 10018.\hskip 1em plus 0.5em
  minus 0.4em\relax Springer, October 2016, pp. 141--152.

\bibitem{KDD.1999}
\BIBentryALTinterwordspacing
I.~U. University~of California. (1999) {KDD} cup 1999 data. [Online].
  Available: \url{http://kdd.ics.uci.edu/databases/kddcup99/kddcup99.html}
\BIBentrySTDinterwordspacing

\bibitem{Tavallaee.2009}
M.~Tavallaee, E.~Bagheri, W.~Lu, and A.~A. Ghorbani, ``A detailed analysis of
  the {KDD CUP 99} data set,'' in \emph{2009 IEEE Symposium on Computational
  Intelligence for Security and Defense Applications}, July 2009, pp. 1--6.

\bibitem{Lemay.2016}
A.~Lemay and J.~M. Fernandez, ``Providing scada network data sets for intrusion
  detection research,'' in \emph{9th Workshop on Cyber Security Experimentation
  and Test (CSET 16)}, Austin, TX, 2016.

\bibitem{Duque_Anton.2018}
S.~Duque~Antón, S.~Kanoor, D.~Fraunholz, and H.~D. Schotten, ``Evaluation of
  machine learning-based anomaly detection algorithms on an industrial
  modbus/tcp data set,'' in \emph{Proceedings of the 13th International
  Conference on Availability, Reliability and Security (ARES)}.\hskip 1em plus
  0.5em minus 0.4em\relax ACM, 2018.

\bibitem{Schneider-Electric.2017}
\BIBentryALTinterwordspacing
S.~Electric. (2017) Life is on. [Online]. Available:
  \url{https://www.schneider-electric.fr/fr/}
\BIBentrySTDinterwordspacing

\bibitem{Drury.2009}
B.~Drury, \emph{Control Techniques Drives and Controls Handbook}, 2nd~ed.

\bibitem{metasploit.}
\BIBentryALTinterwordspacing
Rapid7. metasploit. [Online]. Available: \url{https://www.metasploit.com/}
\BIBentrySTDinterwordspacing

\bibitem{Yeh.2016a}
C.-C.~M. Yeh, Y.~Zhu, L.~Ulanova, N.~Begum, Y.~Ding, H.~A. Dau, D.~F. Silva,
  A.~Mueen, and E.~Keogh, ``Matrix profile i: All pairs similarity joins for
  time series: A unifying view that includes motifs, discords and shapelets,''
  in \emph{2016 IEEE 16th International Conference on Data Mining (ICDM)},
  December 2016, pp. 1317--1322.

\bibitem{Benesty.2009}
J.~Benesty, J.~Chen, Y.~Huang, and I.~Cohen, ``Pearson correlation
  coefficient,'' in \emph{Noise Reduction in Speech Processing}, vol.~2.\hskip
  1em plus 0.5em minus 0.4em\relax Springer, Berlin, Heidelberg, 2009, pp.
  1--4.

\bibitem{Mueen.2010}
A.~Mueen, S.~Nath, and J.~Liu, ``Fast approximate correlation for massive
  time-series data,'' in \emph{Proceedings of the 2010 ACM SIGMOD International
  Conference on Management of Data}, ser. SIGMOD '10.\hskip 1em plus 0.5em
  minus 0.4em\relax New York, NY, USA: ACM, 2010, pp. 171--182.

\bibitem{Mueen.2017}
A.~Mueen, Y.~Zhu, M.~Yeh, K.~Kamgar, K.~Viswanathan, C.~Gupta, and E.~Keogh,
  ``The fastest similarity search algorithm for time series subsequences under
  euclidean distance,'' August 2017,
  \url{http://www.cs.unm.edu/~mueen/FastestSimilaritySearch.html}.

\bibitem{Duque_Anton.2017c}
S.~Duque~Anton, D.~Fraunholz, S.~Teuber, and H.~D. Schotten, ``A question of
  context: Enhancing intrusion detection by providing context information,'' in
  \emph{13th Conference of Telecommunication, Media and Internet
  Techno-Economics (CTTE-17)}, 2017.

\bibitem{Box.2013}
G.~E.~P. Box, G.~M. Jenkins, and G.~C. Reinsel, \emph{Time Series Analysis --
  Forecasting and Control}, 4th~ed., ser. WILEY SERIES IN PROBABILITY AND
  STATISTICS.\hskip 1em plus 0.5em minus 0.4em\relax John Wiley \& Sons, Inc.,
  Hoboken, New Jersey, 2013.

\bibitem{Hochreiter.1997}
S.~Hochreiter and J.~Schmidhuber, ``Long short-term memory,'' \emph{Neural
  Computation}, vol.~9, no.~8, pp. 1735--1780, November 1997.

\bibitem{Gers.1999}
F.~A. Gers, J.~Schmidhuber, and F.~Cummins, ``Learning to forget: Continual
  prediction with lstm,'' \emph{Neural Computation}, vol.~12, pp. 2451--2471,
  1999.

\bibitem{Olah.2015}
\BIBentryALTinterwordspacing
C.~Olah, ``Understanding {LSTM} networks,'' 2015. [Online]. Available:
  \url{http://colah.github.io/posts/2015-08-Understanding-LSTMs/}
\BIBentrySTDinterwordspacing

\bibitem{Chandola.2009}
V.~Chandola, A.~Banerjee, and V.~Kumar, ``Anomaly detection: A survey,''
  \emph{ACM Computing Surveys}, vol.~41, no.~3, pp. 1--58, July 2009.

\bibitem{Yeh.2017}
C.-C.~M. Yeh, N.~Kavantzas, and E.~Keogh, ``Matrix profile vi: Meaningful
  multidimensional motif discovery,'' November 2017, pp. 565--574.

\bibitem{Dau.2017}
H.~A. Dau and E.~Keogh, ``Matrix profile v: A generic technique to incorporate
  domain knowledge into motif discovery,'' in \emph{Proceedings of the 23rd ACM
  SIGKDD International Conference on Knowledge Discovery and Data Mining}, ser.
  KDD '17.\hskip 1em plus 0.5em minus 0.4em\relax New York, NY, USA: ACM, 2017,
  pp. 125--134.

\bibitem{Duque_Anton.2017b}
S.~Duque~Anton, D.~Fraunholz, J.~Zemitis, F.~Pohl, and H.~D. Schotten, ``Highly
  scalable and flexible model for effective aggregation of context-based data
  in generic iiot scenarios,'' in \emph{9th Central European Workshop on
  Services and their Composition. Central European Workshop on Services and
  their Composition (ZEUS-2017), February 13-14, Lugano, Switzerland}, April
  2017, pp. 51--58.

\end{thebibliography}
\end{document}